\pdfoutput=1
\documentclass[letterpaper, 10 pt, conference]{ieeeconf}   

\usepackage{multirow}
\usepackage{graphicx}
\usepackage{booktabs}

\IEEEoverridecommandlockouts                              

\usepackage{graphics} 
\usepackage{epsfig} 
\usepackage{mathptmx} 
\usepackage{times} 
\usepackage{amsmath} 
\usepackage{amssymb}  
\usepackage{hyperref} 
\usepackage[noadjust]{cite}
\usepackage[font={footnotesize}]{caption}

\title{\LARGE \bf
TrACT: A Training Dynamics Aware Contrastive Learning Framework for Long-Tail Trajectory Prediction
}

\author{Junrui Zhang$^{1}$, Mozhgan Pourkeshavarz$^{2*}$, Amir Rasouli$^{2}$\\
\thanks{$^{1}$University of Toronto. Work done during internship at Huawei Technologies Canada. {\tt\footnotesize arielle.zhang@mail.utoronto.ca}}
\thanks{$^{2}$Noah's Ark Laboratory, Huawei Technologies Canada.}
\thanks{*Corresponding author {\tt\footnotesize Mozhgan.Pourkeshavarz@huawei.com}}
}
\begin{document}

\maketitle
\thispagestyle{empty}
\pagestyle{empty}

\begin{abstract}

As a safety critical task, autonomous driving requires accurate predictions of road users' future trajectories for safe motion planning, particularly under challenging conditions. Yet, many recent deep learning methods suffer from a degraded performance on the challenging scenarios, mainly because these scenarios appear less frequently in the training data. To address such a long-tail issue, existing methods force challenging scenarios closer together in the feature space during training to trigger information sharing among them for more robust learning. These methods, however, primarily rely on the motion patterns to characterize scenarios, omitting more informative contextual information, such as interactions and scene layout. We argue that exploiting such information not only improves prediction accuracy but also scene compliance of the generated trajectories. In this paper, we propose to incorporate richer training dynamics information into a prototypical contrastive learning framework. More specifically, we propose a two-stage process. First, we generate rich contextual features using a baseline encoder-decoder framework. These features are split into clusters based on the model's output errors, using the training dynamics information, and a prototype is computed within each cluster. Second, we retrain the model using the prototypes in a contrastive learning framework. We conduct empirical evaluations of our approach using two large-scale naturalistic datasets and show that our method achieves state-of-the-art performance by improving accuracy and scene compliance on the long-tail samples. Furthermore, we perform experiments on a subset of the clusters to highlight the additional benefit of our approach in reducing training bias. 

\end{abstract}

\section{Introduction}

In autonomous driving (AD), future trajectory prediction requires a comprehensive understanding of the driving context, including the agent's history observation and the interactions between the agent and its surroundings. Recent prediction methods \cite{Zhu_2023_CVPR, karim2023destine} rely on such information and show promising results on the existing AD benchmark datasets \cite{caesar2020nuscenes, Chang_2019_CVPR, Sun_2020_CVPR_3}. The performance of these models, however, is not consistent across all scenarios and generally degrades on more challenging ones. One of the reasons for such degradation is the inherent biases of the AD datasets towards simpler scenarios, i.e., lack of challenging cases \cite{chen2023criteria}. As an ill effect, the long-tail issue can pose a safety risk for the development of practical AD systems.

\begin{figure}[t]
\centering
\includegraphics[trim = 8 8 7 8,clip, scale=0.5, width=0.48\textwidth]{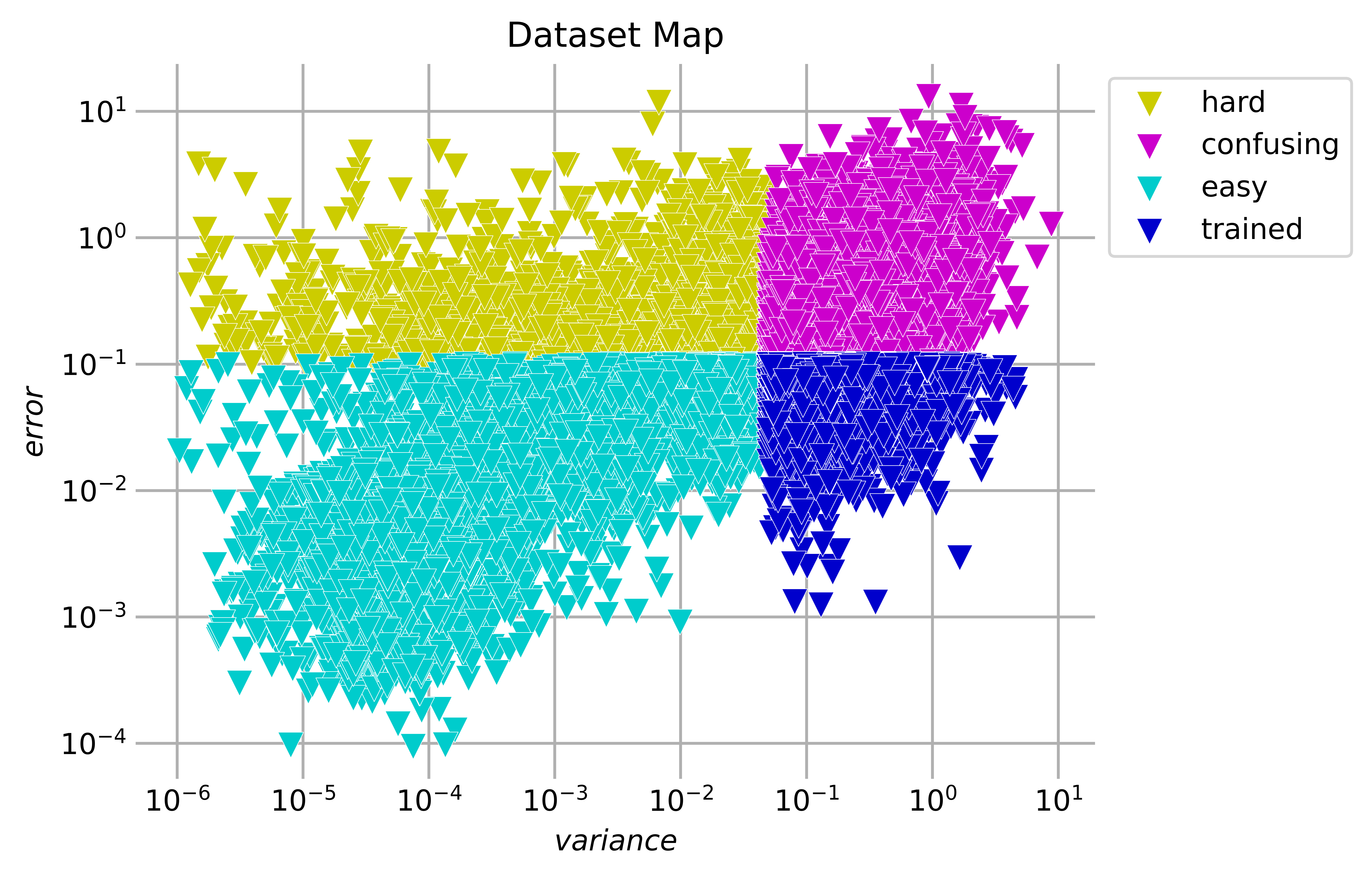}
\caption{A dataset map constructed using the training dynamics information of the baseline model on nuScenes. \(variance\) measures the variance of the samples' errors across all training epochs, and \(error\) measures the samples' errors on the last epoch. The dataset is divided into four clusters, namely \(easy\), \(hard\), \(confusing\), or \(trained\), based on a \(variance\) threshold and an \(error\) threshold. Each sample receives one cluster assignment based on their training behaviors. Only $20\%$ of the data is plotted on the log-scale axis for better visibility.}\label{map}
\vspace{-20pt}
\end{figure}

One way to tackle the long-tail problem is to collect more samples of the challenging scenarios that are infrequent in the dataset. However, this approach is not feasible since such scenarios are generally rare in real-world and mimicking them can pose safety risks. Alternatively, recent approaches \cite{makansi2021exposing, wang2023fend} attempt to solve the long-tail problem during training phase by disentangling the long-tail samples from the rest in the feature space using contrastive learning methods \cite{contestaware}. These works, however, only rely on the motion patterns of the agents, omitting rich contextual information, such as interactions between the agents and their surroundings as well as the scene layout. Such an approach limits the ability of these models to separate features of the challenging samples, hence diminishing their potentials on long-tail predictions.  

In this paper, we propose a training dynamics aware contrastive learning framework, termed TrACT, in which we cluster the model's output errors by exploiting the training dynamics information (see \autoref{map}), namely the prediction errors of the samples at the last epoch and the variance of samples' errors across all training epochs (see \autoref{training}). For a more comprehensive representation of the driving scenario, we use latent features of an encoder-decoder framework that encodes map layout, agents' dynamics, and their interactions with their surroundings. The features are assigned to the clusters based on their output errors' cluster assignments, and are averaged within each cluster to form prototypes. In the next stage, the prototypes are directly passed to the prototypical contrastive learning (PCL) framework for training the final model. 

In summary, our \textbf{main contributions} are as follows: 1) We propose TrACT,  a Training dynamics Aware ContrasTive learning framework, which utilizes the training dynamics information to form data sample clusters with different levels of difficulty on a constructed dataset map. These clusters in conjunction with the feature embeddings of a backbone model form prototypes that are used in a contrastive learning framework to generate robust trajectories. 2) We conduct extensive experimentation on two benchmark datasets for trajectory prediction, namely nuScenes \cite{caesar2020nuscenes} and ETH-UCY \cite{Pellegrini_2009_iccv, Leal_2014_cvpr}, to highlight the effectiveness of TrACT on the top 1-5$\%$ challenging scenarios. 3) We conduct additional studies using safety metrics to show the effect of TrACT on improving scene compliance of the generated trajectories. 4) Lastly, we demonstrate additional benefits of using the dataset map to reduce training bias on the long-tail scenarios.


\section{Related Works}
\subsection{Trajectory Prediction}
Trajectory prediction is a fundamental problem in autonomous driving. Accurate forecasting requires an understanding of the surrounding agents and the interactions among them as well as the scene configuration. To effectively process such multimodal information, existing methods rely on various architectures, including recurrent networks \cite{wang2023fend, Rasouli_2021_ICCV, crowdinteraction}, CNNs \cite{gilles2022gohome,ye2021tpcn}, GNNs \cite{mohamed2020social, salzmann2020trajectron++, li2020evolvegraph, casas2019spatially, liang2020learning, jia2023hdgt, Pourkeshavarz_2023_ICCV}, and more recently transformers \cite{rasouli_2023_icra, Shi_2023_ICCV, karim2023destine,amirloo2022latentformer}. One of the key challenges in prediction is the uncertainty of future behaviors. To address this problem, models resort to generating a diverse set of trajectories using different approaches \cite{Mao_2023_CVPR, Jiang_2023_CVPR, Nikdel_2023_ICRA, Wang_2023_ICRA_1}, one of which is CVAE \cite{rasouli2023novel, Lee_2022_CVPR, yuan2021agentformer, salzmann2020trajectron++}. We use \cite{makansi2021exposing }, which is a variant of a CAVE-based model \cite{salzmann2020trajectron++} as the backbone.

\subsection{Long-tail Learning}
Trajectory prediction models are often evaluated on large-scale datasets \cite{caesar2020nuscenes, Chang_2019_CVPR, Sun_2020_CVPR_3, Leal_2014_cvpr} and the performance is averaged over all samples in the data. Despite achieving promising results on the benchmarks, models trained on these datasets underperform on challenging scenarios \cite{Pourkeshavarz_2023_ICCV, makansi2021exposing, wang2023fend}.  Such an issue is largely due to the long-tail nature of these datasets as they are more biased towards the common scenarios and contain much smaller number of challenging cases \cite{chen2023criteria}.

The long-tail phenomenon is incurred by the imbalanced number of more frequent samples and less frequent samples in the dataset. There are many studies on improving the long-tail learning on classification tasks offering techniques, such as data resampling \cite{chawla2002smote, Han_2005_AIC, shen2016relay}, loss reweighting \cite{cui2019class, He2009LearningFI, lin2017focal}, boundary adjustment \cite{cao2019learning, menon2020long}, and more recently feature and label distribution smoothing \cite{yang2021delving}.

Some recent works address the long-term problem in trajectory prediction \cite{wang2023fend, makansi2021exposing}. The authors of \cite{makansi2021exposing} learn long-tail samples  by reshaping the feature space to better distinguish the head and tail samples through contrastive learning. The work in  \cite{wang2023fend} includes an additional step of offline clustering to obtain the pseudo labels for the prototypical contrastive learning framework. However, in \cite{makansi2021exposing, wang2023fend}, the authors argue that tail samples contain more evasive maneuvers, resulting in more complicated trajectory shapes. Therefore the learning is purely based on the motion patterns of individual agents. Here, however, crucial contextual information, such as interactions and scene layouts that are necessary for characterizing motions are omitted. Moreover, for contrastive learning, these approaches use heuristic methods for clustering challenging scenarios using the shape of the observed trajectories. These methods are not effective as they do not examine sample difficulty on the scenario level. In this work, we examine the model learning capability of each sample to determine the sample difficulty by exploiting the training dynamics information and building a dataset map similar to \cite{swayamdipta2020dataset} to classify data samples into four different clusters for the prototypical contrastive learning.

\begin{figure}[t]
\centering
\includegraphics[trim=8 8 6 6, clip,width=0.48\textwidth]{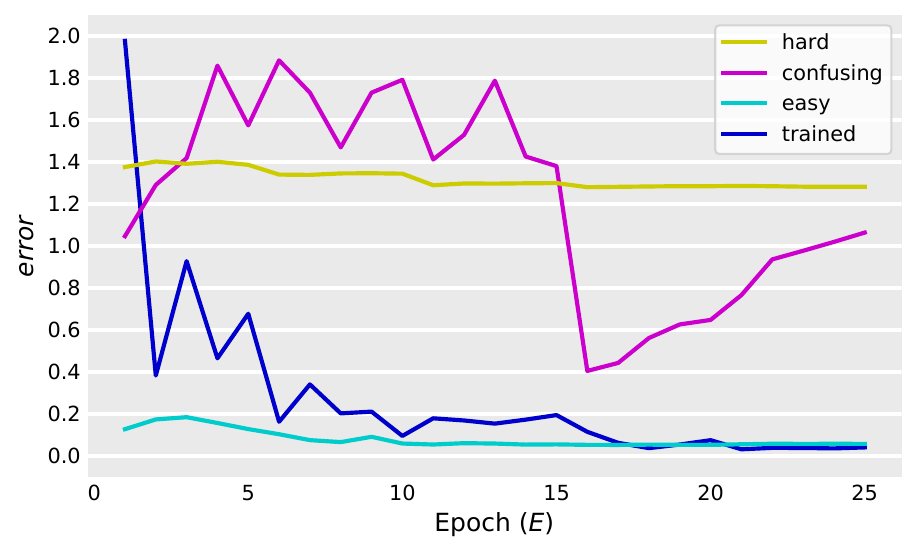}
\caption{Error evolution during training for one sample from each one of the \(hard\), \(confusing\), \(easy\), and \(trained\) clusters on nuScenes.
}
\label{training}
\vspace{-20pt}
\end{figure}

\begin{figure*}[t]
\centering
\includegraphics[trim=0 90 0 0 , clip,width=\textwidth]{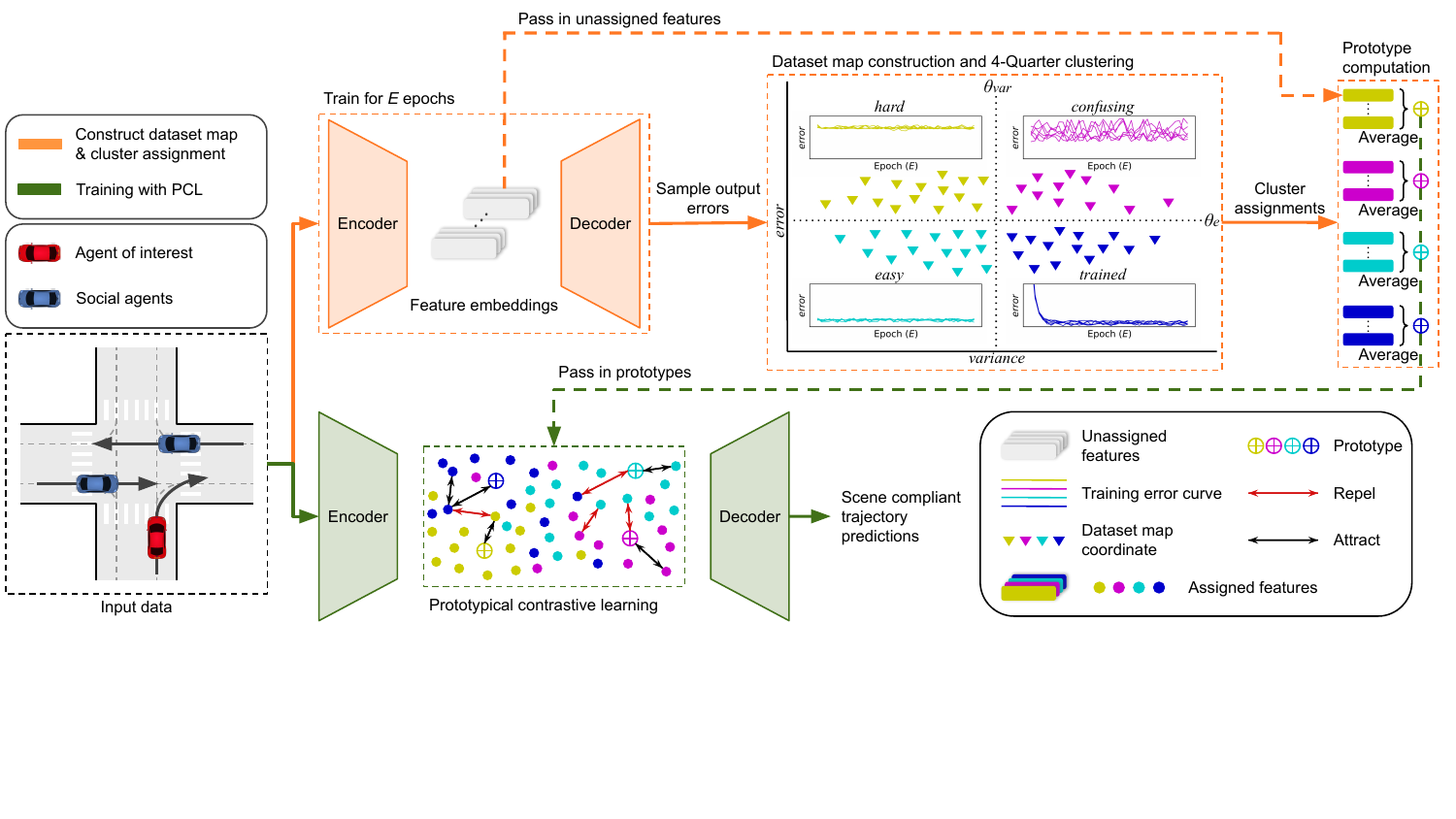}
\caption{An overview of the training dynamics-aware contrastive learning framework. The orange module represents the first training phase to construct the dataset map and performs the cluster assignment of each sample for prototype computation. The green module represents the second training phase with the prototypical contrastive loss.}
\label{pipelilne}
\end{figure*}


\section{Method}

\subsection{Problem Formulation}
The task of trajectory prediction is to predict agents' future states given their observed history. Mathematically, for the $i$-th agent at time step $t$, let consider the past trajectory of the $i$-th agent as a set of 2D coordinates in bird's eye view over some observation horizon $L$ time steps $X_i = \{(x_i, y_i)^{t-L}, \cdots,(x_i, y_i)^{t-1}\}$. Then, the goal is to predict future trajectories $Y_i = \{(x_i, y_i)^{t+1}, \cdots,(x_i, y_i)^{t+T}\}$, where $T$ is the prediction horizon. Additionally, the driving map configuration is provided in the form of an HD map.

\subsection{Overview}

As shown in \autoref{pipelilne}, we propose a training dynamics aware contrastive learning framework, which consists of an offline clustering module using the training dynamics information and a prototypical contrastive learning module. We first train the model to obtain the training dynamics information from sample errors, and then construct the dataset map accordingly. Next, we segment the dataset map into clusters containing samples with similar difficulties and compute the prototypes by averaging the features belonging to the same clusters. Lastly, with the prototypes, we train the model with PCL to generate the final predictions.

\subsection{Constructing Dataset Map}
To construct the dataset map, we train the model on the whole training dataset and record the error evolution during training of each individual sample. In general, there are four types of training behaviors and for each sample \(i\) such behaviors can be described effectively by two variables: the last epoch error denoted as \(error_i\), and the variance of the errors across all training epochs \(E\) denoted as \(variance_i\). Here, \(error_i\) informs the predictability of each sample of the converged model and \(variance_i\) informs the learning difficulty of each sample since the change in the error during training reflects the model knowledge gain, i.e. good-fit or underfit. Therefore both metrics can be used for the assessment of the sample's difficulty. Subsequently, we define the training dynamics information of each sample to be the coordinates on the dataset map, with x coordinate corresponding to \(variance_i\) and y coordinate corresponding to \(error_i\). The dataset map provides a proper segmentation for forming clusters of samples with different levels of difficulty. 

\subsection{Clustering}
By analyzing the training behavior of each individual sample, we discover that in a dataset, samples that have low \(error\)s may have different \(variance\)s which shows whether the model parameters are being updated correctly during training in order to learn these samples. This also holds for high \(error\)s, hence taking both dimensions into consideration, we propose a \textbf{4-Quarter} clustering method.

\textbf{4-Quarter.} Based on different training behaviors, we first segment the dataset map into four quarters with different levels of difficulty using an error threshold \(\theta_{e}\) and a variance threshold \(\theta_{var}\). Accordingly, the samples from the same quarter form a cluster. The four different training behaviors for the clusters, as shown in \autoref{training}, are as follows: the \(hard\) cluster (a high \(error\) but a low \(variance\)) indicates that the model achieves a small improvement over the training period; The \(confusing\) cluster (a high \(error\) and a high \(variance\)) indicates that the model's performance is highly fluctuating during training, thus the model is confused; The \(easy\) cluster (a low \(error\) and a low \(variance\)) indicates that the model consistently predicts well during training; And the \(trained\) cluster (a low \(error\) and a high \(variance\)) indicates that the model is improving over time.

\subsection{Prototype Computation}
After clustering, we compute one single prototype for each cluster for the PCL learning. We first extract all the unassigned features from the encoder feature space and based on the dataset map cluster assignments, we take the average of all features in the same cluster to form the prototype. Afterwards, we pass the prototypes to the prototypical contrastive learning module to better organize the feature space for the encoder.

\subsection{Single Level Prototypical Contrastive Learning}
\label{sec:PCL}
The original PCL is an estimation-expectation (EM) algorithm which re-estimates the cluster label at each iteration \cite{li2020prototypical}. We replace the estimation step with the offline clustering to obtain the prototypes. Following \cite{wang2023fend}, we apply the PCL loss to the bottleneck of the encoder-decoder architecture but with a single level of clusters. In the feature space, the PCL loss will group the same cluster samples together to encourage the model to distinguish between samples with different levels of difficulties. The loss consists of two terms:
\begin{equation}
    L_{ProtoNCE} = L_{ins} + L_{proto},
    \label{eq1}
\end{equation}
where \(L_{ins}\) is the instance-wise term that penalizes large distances of the same cluster samples in the feature space and  \(L_{proto}\) is the instance-prototype term that penalizes large distances between the sample and its cluster prototype in the feature space.

\textbf{Instance-wise term.}
The instance-wise loss term \(L_{ins}\) in \autoref{eq1} is designed to speed up convergence by attracting instances belonging to the same cluster:
\begin{equation}
    L_{ins} = -\sum\limits_{i=1}^{r}\frac{1}{N_{po_i}}\sum\limits_{i_+=1}^{N_{po_i}}log\frac{exp(v_i \cdot v_{i_+}/\tau)}{\Sigma_{j=1}^rexp(v_i \cdot v_j / \tau)},
    \label{eq2}
\end{equation}
where \(r\) denotes the batch size, \(N_{po_i}\) is the number of same-cluster samples \(i_+\)s of an arbitrary sample \(i\), and \(v\) is the feature embedding. \(j \in [1, ..., r]\) represents all available samples in the batch and \(\tau\) is the contrastive temperature.

\textbf{Instance-prototype term.}
In our method, we put more emphasis on the overall difficulty of each training sample instead of solely the motion patterns. Hence, we do not need a hierarchy of clusters to accommodate for granularity. Here, unlike the original PCL loss \cite{li2020prototypical}, our implementation of the \(L_{proto}\) term in \autoref{eq1} only has one level of cluster:
\begin{equation}
    L_{proto} = -  \sum_{i=1}^{r} log \frac{exp(v_i \cdot c_i / \phi_i)}{\sum_{j=1}^{4}exp(v_i \cdot c_j / \phi_j)},
    \label{eq3}
\end{equation}
where for an arbitrary sample \(i\), \(c\) is the cluster prototype, \(\phi\) is the cluster density, and \(j \in [1, ..., 4]\) represents all four clusters. The density \(\phi\) of the cluster is given by,
\begin{equation}
    \phi = \frac{\sum_{z=1}^{Z}||v_z-c_z||_2}{Zlog(Z+\alpha)}, \label{eq4}
\end{equation}
where $Z$ is the number of samples belonging to the cluster, \(z\) is an arbitrary sample from the cluster, and \(\alpha\) is a smoothing term, which is set to $10$ following \cite{li2020prototypical}. 

In the end, our final training objective is described as,
\begin{equation}\label{eq5}
    L = L_{reg} + \lambda L_{ProtoNCE},
\end{equation}
where \(\lambda\) is the control weight of the prototypical loss.


\begin{table*}[t]
\centering
\caption{The results (minADE minFDE KDE-NLL) on the nuScenes and ETH-UCY datasets. For all values, lower is better. The "*" represents the reported results from FEND \cite{wang2023fend} on nuScenes.}
\label{maintable}
\addtolength{\tabcolsep}{-4pt}
\renewcommand{\arraystretch}{1.2} 
\resizebox{\textwidth}{!}{%
\begin{tabular}{l|l|ccc|ccc|ccc|ccc|ccc|ccc}
\specialrule{1.5pt}{1pt}{0pt}
\textbf{Dataset}&\textbf{Method} & \multicolumn{3}{c|}{\textbf{Top 1$\%$}}  & \multicolumn{3}{c|}{\textbf{Top 2$\%$}} & \multicolumn{3}{c|}{\textbf{Top 3$\%$}} & \multicolumn{3}{c|}{\textbf{Top 4$\%$}} & \multicolumn{3}{c|}{\textbf{Top 5$\%$}} & \multicolumn{3}{c}{\textbf{All}} \\ 
\hline
\multirow{6}{*}{\textbf{nuScenes}}&Traj++ EWTA$^*$ & 1.33&3.09& - & 1.02&2.35&-& 0.87&2.00&- & 0.80&1.80&- &  0.74&1.64&- & 0.19&0.32&- \\ 
\cline{2-20}
\cline{2-20}
& +contrastive$^*$ & 1.28&2.85&- & 0.97&2.15&- & 0.83&1.83&- & 0.76&1.64&- & 0.70&1.48& -& \textbf{0.18}&\textbf{0.30}&-\\
& +FEND$^*$& 1.21&2.50&-&0.92&1.88&-&0.79&1.61&-&0.72&1.43&-&0.66&1.31&-&0.17&0.26&-\\ 
\cline{2-20}
&Traj++ EWTA & 1.73&4.43&11.72 & 1.36&3.54&10.02 & 1.17&3.03&8.80 & 1.04&2.68&7.83 &  0.95&2.41&7.21 & 0.19&0.32&-0.14 \\ 
\cline{2-20}
\cline{2-20}
& +contrastive& 1.33&3.09&8.91 & 1.04&2.44&7.81 & 0.90&2.08&7.05 & 0.81&1.85&6.37 & 0.75&1.68&5.96 & \textbf{0.18}&\textbf{0.30}&-0.11 \\
& \textbf{+TrACT (ours)} & \textbf{1.23}&\textbf{2.65}&\textbf{7.22} & \textbf{0.98}&\textbf{2.11}&\textbf{6.27} & \textbf{0.85}&\textbf{1.82}&\textbf{5.54} & \textbf{0.78}&\textbf{1.64}&\textbf{4.98} & \textbf{0.72}&\textbf{1.49}&\textbf{4.62} & 0.19& 0.31 &\textbf{-0.21}  \\
\hline
\hline
\multirow{3}{*}{\textbf{ETH-UCY}}&Traj++ EWTA & 0.98&2.54&8.71 & 0.79&2.07&5.45 & 0.71&1.81&4.89 & 0.65&1.63&4.10 & 0.60&1.50&3.53 & \textbf{0.17}&\textbf{0.32}&\textbf{-0.42} \\
\cline{2-20}
\cline{2-20}
 &+contrastive & 0.92&2.33& 7.85& 0.74&1.91&5.02  & 0.67&1.71& 4.54& 0.60&1.48&3.71 & 0.55&1.32& 3.13& \textbf{0.17}&\textbf{0.32}&-0.22 \\
&+FEND & 0.84&2.13& -& 0.68&1.68&-  & \textbf{0.61}&\textbf{1.46}& -& \textbf{0.56}&\textbf{1.30}&- & \textbf{0.52}&1.19 &-& \textbf{0.17}&\textbf{0.32}&- \\
&\textbf{+TrACT (ours)} 
&\textbf{0.80} &\textbf{2.00}&\textbf{3.39} &\textbf{0.65} &\textbf{1.63}&\textbf{2.52} &\textbf{0.61} &\textbf{1.46}& \textbf{2.34}
&\textbf{0.56} &1.31&\textbf{2.11} &\textbf{0.52} &\textbf{1.18}&\textbf{1.93} &\textbf{0.17} &\textbf{0.32}& -0.25\\
\specialrule{1.5pt}{0pt}{1pt}
\end{tabular}%
}
\end{table*}

\begin{table*}[t]
\centering

\renewcommand{\arraystretch}{1.2} 
\addtolength{\tabcolsep}{-4pt}
\caption{The results (HOR SOR) in percentage (\%) on nuScenes for scene compliance analysis. For all values, lower is better.}
\label{viol}
\resizebox{0.6\textwidth}{!}{
\begin{tabular}{l|c|c|c|c|c|c}
\specialrule{1.5pt}{1pt}{0pt}
\multirow{2}{*}{\textbf{Method}}&\textbf{Top}&\textbf{Top}&\textbf{Top}&\textbf{Top}&\textbf{Top}&\multirow{2}{*}{\textbf{All}}\\
&\textbf{1\%} & \textbf{2\%} & \textbf{3\%} & \textbf{4\%} & \textbf{5\%} &  \\ \hline
Traj++ EWTA&6.52 1.22&5.75 1.05&4.24 0.76&3.80 0.61&3.17 0.49&0.22 0.03\\ 
\cline{1-7} \cline{1-7}
+contrastive &4.65 0.72&3.42 0.65&2.69 0.50& \textbf{2.17} 0.38 & \textbf{1.93} 0.33 & \textbf{0.18 0.02} \\
\textbf{+TrACT (ours)} &\textbf{4.04 0.56}&\textbf{3.26 0.56}&\textbf{2.59 0.42}& 2.41 \textbf{0.36}& 1.99 \textbf{0.29}& 0.23 0.03\\ 
\specialrule{1.5pt}{0pt}{1pt}
\end{tabular}}\vspace{-3mm}
\end{table*}

\section{Experiments}

\subsection{Setup}

\textbf{Datasets.} We evaluate TrACT on two benchmark trajectory prediction datasets: nuScenes \cite{caesar2020nuscenes} and ETH-UCY \cite{Pellegrini_2009_iccv,Leal_2014_cvpr}. nuScenes is a large-scale autonomous driving dataset containing 1000 scenes, which are split into 850 and 150 scenes for training-validation and testing, respectively. In our experiments, we used the preprocessed Trajectron++ nuScenes dataset, which excludes 3 scenes from the training-validation split and further divides the split into 697 scenes for training and 150 scenes for validation. In other words, on this dataset, we conduct all our evaluations on the validation set. In nuScenes, each scene contains a 20-second driving scenario with corresponding HD maps. Following \cite{wang2023fend, makansi2021exposing} we focus on the vehicles in nuScenes.

ETH and UCY are pedestrian datasets with five subsets, including ETH, Hotel, Univ, Zara1, and Zara2, all containing different number of scenes for training and testing. Each scene contains a recording of pedestrians interacting scenario with varying time length. Following \cite{Shi_2023_ICCV, makansi2021exposing, wang2023fend}, we perform 5-fold cross validation on the five subsets.

\textbf{Metrics.} We use common evaluation metrics \cite{Shi_2023_ICCV,salzmann2020trajectron++, wang2023fend}: Average Displacement Error (ADE), which is the average \(L_2\) distance between all predicted states and the ground truth and Final Displacement error (FDE), which is the \(L_2\) distance between the last predicted state and the ground truth. Following \cite{salzmann2020trajectron++}, we also report on Kernel Density Estimate-based Negative Log Likelihood (KDE-NLL) which is the average negative log of the ground truth probability density under the distribution created by fitting a kernel density estimate on top \(K\) predicted candidates. For distance-based metrics, we report the results on best-of-20 denoted by minADE/minFDE, and for KDE-NLL \(K=20\) as well.

To evaluate the map compliance of the predicted trajectories, we also report on two safety metrics adopted from \cite{bahari2022vehicle}: Hard Off-Road Rate (HOR) which is the percentage of test samples that have at least one predicted point appearing off-Road; Soft Off-road Rate (SOR) which is the percentage of off-road predicted points with respect to all predicted points.

\textbf{Models.} We use Trajectron++ EWTA (Traj++ EWTA) \cite{makansi2021exposing} as our baseline, which is an improved variation of Trajectron++ \cite{salzmann2020trajectron++}. Traj++ EWTA replaces the conditional variational autoencoder in the original model with a multi-hypothese decoder following an evolving-winner-takes-all (EWTA) training schedule. We also compare TrACT with state-of-the-art FEND (ICCV2023) \cite{wang2023fend} and Traj++ EWTA +contrastive (ICCV2021) \cite{makansi2021exposing} methods, which, similar to TrACT, are specifically designed to address the long-tail problem in trajectory prediction.

\textbf{Long-tail samples selection.} To evaluate TrACT's performance on long-tail samples, following \cite{wang2023fend}, we first evaluate baseline Traj++ EWTA model on test sets and then select the top 1-5$\%$ challenging samples with the largest minFDEs as the five challenging subsets. minFDE is an effective metric for long-tail samples because it focuses on the accuracy of endpoint (target) prediction, which is more crucial for planning tasks. We also use minFDE for the measurement of \(error\) to collect the training dynamics information.

\textbf{Implementation details.}
\label{implementation}
For learning training dynamics information, we followed the training schedule method in \cite{makansi2021exposing}, trained on Traj++ EWTA with a batch size of 256 and an EWTA schedule of top$_{ \{20,10,5,2,1\} }$ with 5 and 100 epochs for each EWTA stage for nuScenes and ETH-UCY, respectively.

To track the training behaviors, we saved the minFDEs of each sample at every epoch on nuScenes and every 20 epoch on ETH-UCY, as well as the feature embeddings \(v_i\) of each sample at the final epoch. The feature embeddings by default have a size of 232 for pedestrians and a size of 264 for vehicles \cite{makansi2021exposing}. Same as \cite{salzmann2020trajectron++}, we set \((L, T)\) to \((8, 6)\) and \((7, 12)\) for nuScenes and ETH-UCY, respectively. The duration for each timestep is set to be $0.5$s for nuScenes and $0.4$s for ETH-UCY.

For the contrastive learning step, we used the same training schedule as in \cite{makansi2021exposing} but trained with 10 and 100 epochs for each EWTA stage for nuScenes and ETH-UCY, respectively. For the single cluster PCL loss, we empirically set \(\lambda=0.01\), \(\tau=0.1\) for nuScenes, and \(\lambda=10\), \(\tau=0.5\) for ETH-UCY. Finally, we set thresholds \{\(\theta_{err}=0.70\), \(\theta_{var}=0.20\)\} for nuScenes, and \{\(\theta_{err}=0.70\), \(\theta_{var}=0.15\)\} for ETH-UCY. To avoid numerical overflow, we scaled the cluster density \(\phi_i\) for each sample \(i\) by $100$ for ETH-UCY.

\subsection{Comparison to SOTA}
We compare TrACT against baseline Traj++ EWTA, Traj++ EWTA + contrastive, \cite{makansi2021exposing} and state-of-the-art FEND \cite{wang2023fend} methods on nuScenes and ETH-UCY and report the results in \autoref{maintable}. Note that, on nuScenes, two sets of baseline results are reported: the original ones from FEND \cite{wang2023fend} ($^*$) and the ones we reproduced following instructions in FEND. While we reproduced the same baseline results on ETH-UCY following instructions in FEND, we observed a large discrepancy on the baseline results on nuScenes. As a result, we primarily compare to the reproduced baseline performance on nuScenes and report results from FEND as a reference. FEND does not report results on NLL-KDE and in the absence of publicly available code, we were not able to evaluate FEND on this metric. As shown in \autoref{maintable}, TrACT achieves state-of-the-art performance on all metrics on all challenging subsets, significantly improving performance by up to $22.48\%$ on KDE-NLL on the top $5\%$ challenging subset compared to Traj++ EWTA + constrastive. On distance-based metrics, the largest improvement is achieved on the top 1$\%$ challenging subset with $7.52\%$ on minADE and $14.24\%$ on minFDE. On all samples, TrACT performs the second best with a very small margin on distance-based metrics while maintaining the best performance on KDE-NLL.

Similarly, on the ETH-UCY dataset, TrACT  achieves state-of-the-art performance on most metrics on all challenging subsets and achieves the second best minADE on the top $4\%$ most challenging subset with a small margin. Here, once again TrACT achieves the largest improvement of 4.76$\%$ for minADE, and 6.10$\%$ for minFDE on the top $1\%$ challenging subset compared to FEND. Contrary to nuScenes, the best improvement of 56.82$\%$ for KDE-NLL is achieved on the top 1$\%$ challenging subset compared to Traj++ EWTA +contrastive. For all samples, TrACT maintains the best performance for distance-based metrics but performs second best on KDE-NLL. Such a difference on performance can be attributed to the differences between the two datasets. In the absence of scene layout information, on ETH-UCY, the model relies more on the motion and interaction information.

\subsection{Scene Compliance Predictions on Challenging Scenarios}
We evaluate TrACT's ability to generate scene compliant trajectories. We use HOR and SOR metrics to compare TrACT with the baseline and its variation with contrastive learning on nuScenes. As shown in \autoref{viol}, TrACT exhibits a significant improvement across most challenging subsets on both metrics by up to $13.11\%$ and $22.22\%$ on HOR and SOR, respectively while achieving a close second on SOR on the top $4\%$ and top $5\%$ challenging subsets.  Although the performance on all samples for SOR and HOR are slightly affected by a small drop, which can be attributed to the trade-off introduced by the contrasting learning, the improvements on the challenging subsets highlight the advantage of TrACT, which exploits rich contextual information for clustering as part of the contrastive learning. 


\begin{figure}[t]
\centering 
\includegraphics[width=0.48\textwidth]{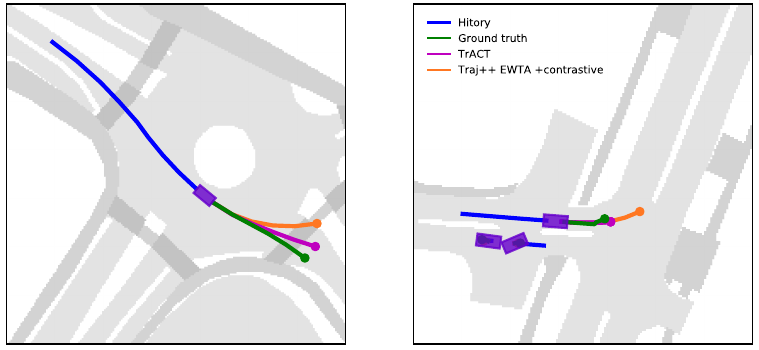}
\caption{Qualitative results on the nuScenes dataset: left shows a complicated scene layout in which the baseline produces an off-road prediction, whereas TrACT does not; Right shows TrACT producing lane compliance predictions compared to the baseline.
}
\label{qualitative}
\vspace{-20pt}
\end{figure}


\begin{table*}[t]
\addtolength{\tabcolsep}{-4pt}
\renewcommand{\arraystretch}{1.2} 
\caption{Ablation study on \(\theta_e\) and \(\theta_{var}\) showing the percentage of each cluster and the results in the format of (minADE minFDE KDE-NLL) of top 1$\%$ and top 5$\%$ challenging samples as well as all samples}
\centering\label{thresholds}
\resizebox{0.8\textwidth}{!}{
\begin{tabular}{l|c|c|c|c|c|c|c}
\hline
\textbf{\(\{\theta_{e}, \theta_{var}\}\)} &
\multicolumn{1}{c|}{\textbf{Easy (\%)}} & \textbf{Confusing (\%)} & \textbf{Hard (\%)} & \multicolumn{1}{c|}{\textbf{Trained (\%)}} &
\textbf{Top 1\%} & \textbf{Top 5\%} & \textbf{All} \\ \hline
\{0.50, 0.10\} & 55.72 & 20.54 & 4.43 & 19.30 
& 1.30 \textbf{2.50} 7.35 & 0.84 1.51 4.56 & 0.26 0.39 0.82 \\ \hline
\{0.70, 0.20\} & 65.53 & 11.06 & 3.68 & 19.71 
& \textbf{1.23} 2.65 7.22 & \textbf{0.72} 1.49 4.62 & \textbf{0.19 0.31 -0.21} \\ \hline
\{0.90, 0.10\} & 59.16 & 8.01 & 1.00 & 31.83 
& 1.25 2.54 \textbf{7.16} & 0.76 \textbf{1.46 4.21} & 0.22 0.35 0.27 \\ \hline
\end{tabular}}\vspace{-3mm}
\end{table*}

\begin{table*}[t]
\addtolength{\tabcolsep}{-4pt}
\renewcommand{\arraystretch}{1.2} 
\centering
\caption{Traj++ EWTA results (minADE minFDE NLL-KDE) on the full nuScenes dataset and nuScenes with 20$\%$ of the \(easy\) samples removed}
\label{clusters}
\resizebox{0.8\textwidth}{!}{
\begin{tabular}{l|ccc|ccc|ccc|ccc|ccc|ccc}
\hline
\textbf{Training Dataset (size$\%$)}&\multicolumn{3}{c|}{\textbf{Top 1$\%$}} &\multicolumn{3}{c|}{\textbf{Top 2$\%$}} &\multicolumn{3}{c|}{\textbf{Top 3$\%$}} &\multicolumn{3}{c|}{\textbf{Top 4$\%$}}
&\multicolumn{3}{c|}{\textbf{Top 5$\%$}} 
&\multicolumn{3}{c}{\textbf{All}} \\ \hline
Full Dataset (100$\%$) & 1.73&4.43&11.72 & 1.36&3.54&10.02 & 1.17&3.03&8.80 & 1.04&2.68&7.83 &  0.95&2.41&7.21 & 0.19&0.32&\textbf{-0.14} \\ \hline
Removed easy ($\approx80\%$) &\textbf{1.37}&\textbf{3.10}& \textbf{9.11} &\textbf{1.06}&\textbf{2.42}& \textbf{8.05} &\textbf{0.92}&\textbf{2.08}& \textbf{7.27}&\textbf{0.83}&\textbf{1.83}&\textbf{6.63}& \textbf{0.76}& \textbf{1.65}&\textbf{6.26}& \textbf{0.19}&\textbf{0.32}&0.18\\ \hline
\end{tabular}
}\vspace{-4mm}
\end{table*}
\subsection{Qualitative Results} In \autoref{qualitative}, we present the qualitative examples of TrACT in comparison to the baseline. The left plot shows TrACT's ability to reason through challenging map layouts, such as roundabouts, producing a more compliant prediction that does not go off-road, whereas Traj++ EWTA + contrastive failed to do so. The right plot shows that TrACT can determine the correct lane direction and avoid generating trajectory that turns into the opposite lane while the baseline method fails to do so. In both cases, the motion pattern is simple but the contextual information is more complicated, requiring the model to rely more on the map information to make safe and scene compliant predictions.

\subsection{Ablation Studies}
\begin{figure}[t]
\centering
\includegraphics[scale=0.48]{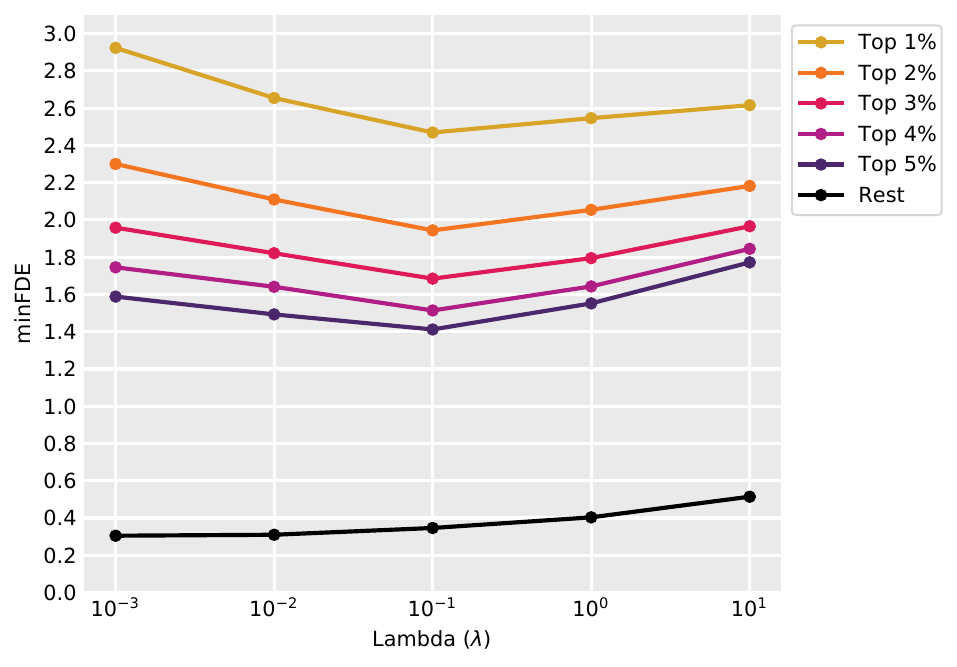}
\caption{Parameter sensitivity of \(\lambda\) on minFDE, plotted for the top 1-5$\%$ challenging subsets and the rest 95$\%$ of the data.}\vspace{-3mm}
\label{lambda}
\end{figure}

\textbf{Control of the contrastive loss.}
In \autoref{eq5}, we introduced \(\lambda\) to control the contrastive loss term. For this study, we examine parameter sensitivity with respect to \(\lambda\) and illustrate the results on log-scale in \autoref{lambda}. As \(\lambda\) increases, the overall loss is more dominated by the contrastive loss, causing a decrease in minFDE for the top 1-5$\%$ challenging subsets. However, there is a trade-off between the performance on the tail samples and the head, i.e., the rest of the samples (shown as 'Rest'). The increasing trend of minFDE on the head samples indicates a declining  performance as \(\lambda\) increases.

\textbf{Choice of \(\theta_e\) and \(\theta_{var}\).} In the clustering step,  \(\theta_e\) and \(\theta_{var}\) thresholds determine the size of each cluster. To study the impact of these thresholds, we conducted a grid search with different \(\theta_e\) and \(\theta_{var}\) values. Based on the statistics of the \(error\)s and \(variance\)s, we empirically chose the search range for both thresholds: \(\theta_{e} \in [0.30, 0.50, 0.70, 0.90]\) and \(\theta_{var} \in [0.05, 0.10, 0.15, 0.20]\) to cover a variety of different cluster compositions. We discovered that when the \(easy\) cluster is in range \(55\%\)-\(65\%\) the model achieves the best performance. Therefore, in \autoref{thresholds}, we show the performance change on three metrics minADE, minFDE, and KDE-NLL on nuScenes, with different combinations of the thresholds, including 55\%, 60\%, 65\% of the \(easy\) cluster.

The combination \(\{\theta_{e}=0.70, \theta_{var}=0.20\}\) maintains the best balance between the performance on the top 1$\%$ and top 5$\%$ challenging subsets, and all samples. This combination emphasizes more on the non-\(easy\) clusters. Similarly, we made the choice of \(\{\theta_{e}=0.70, \theta_{var}=0.15\}\) for ETH-UCY.

\subsection{Dataset Map for Reducing Training Bias}
In this study, we aim to show the benefit of the proposed dataset map for achieving better performance without contrastive learning. For this purpose, we conducted experiments on the baseline model, Traj++ EWTA. We train the model on the full $100\%$ dataset and also on $80\%$ of the data by randomly removing $20\%$ of the data that belong to the \(easy\) cluster. Our intuition is that by removing a portion of the \(easy\) samples, we reduce the overall data bias, as the \(easy\) samples are more frequent in the dataset. Hence, the model would focus more on challenging scenarios, and as a result, achieves a more balanced performance without the use of an explicit contrastive objective. As our findings in \autoref{clusters} suggest, improvements of up to $31.72\%$ on distanced-based metrics and $22.27\%$ on KDE-NLL are achieved across all challenging subsets while the overall performance is unchanged on distance-based metrics. The decline in KDE-NLL metric on all samples can be primarily due to the reduced size of the dataset making the model unsure about the true distributions of the samples.  

\section{Conclusions}
In this work, we proposed a novel framework for learning long-tail scenarios in the context of trajectory prediction for autonomous driving. Our approach, TrACT, exploits the training dynamics information of the model to cluster samples into groups of different levels of difficulty. The clusters, combined with the model feature embeddings, form prototypes to be used in a prototypical contrastive learning framework.

We conducted empirical studies on two trajectory prediction benchmark datasets and showed that TrACT achieved state-of-the-art performance by significantly improving over past arts across the challenging subsets. Besides achieving improved performance on common metrics, TrACT generates significantly more map compliant trajectories, making it more suitable for practical applications. At the end, we illustrated the benefit of the proposed dataset map construction technique for improving performance on challenging scenarios without explicit use of a contrastive learning objective.

\bibliographystyle{IEEEtran}
\bibliography{./IEEEabrv, ./IEEEexample}




\end{document}